\documentclass[letterpaper, 10 pt, conference]{ieeeconf}  
\usepackage{url}
\usepackage{float}
\usepackage{framed}
\usepackage{amsmath}
\usepackage{amssymb}
\usepackage{graphicx}
\usepackage{subcaption}
\usepackage{lipsum}
\usepackage{caption}
\usepackage[export]{adjustbox}

\IEEEoverridecommandlockouts                              

\title{\LARGE \bf
Radiation Surveys in Active Nuclear Facilities with Heterogeneous Collaborative Mobile Robots
}

\author{
    Mitchell Pryor, Alex Navarro, Janak Panthi, Kevin Torres, Mary Tebben, \\  Daniel Meza, Caleb Horan, Alex Macris\\
    \textit{The University of Texas at Austin, Nuclear and Applied Robotics Group}
} %

\begin{document}

\maketitle
\thispagestyle{empty}
\pagestyle{empty}

\begin{abstract}

Nuclear facilities must routinely survey their infrastructure for radiation contamination. Generally, this is done by trained professionals, wearing personal protective equipment (PPE) that swipe potentially contaminated surfaces and test the wipes under detectors. This approach leaves personnel vulnerable to radiation exposure and is not comprehensive. Robots address these inadequacies, offering a cost-effective solution with negligible downtime. We present a Robot Radiation Survey System (RRSS): a heterogeneous robot team to perform comprehensive alpha/beta/gamma radiation surveys. The RRSS system members, core capabilities, and comprehensive survey plan are addresses in this paper. 

\end{abstract}

\section{Introduction}
\label{sec:introduction}

Efficient and effective monitoring of radiation levels is crucial for the long-term maintenance of nuclear facilities and the safety of nuclear workers. Gamma and beta radiation are detectable some distance from the radiation source, facilitating the use of non-contact radiation sensors for routine surveys. Alpha radiation, on the other hand, is undetectable more than a few centimeters from the sensor. Alpha detectors operate within a small, constant offset measuring just a handful of millimeters from the source to obtain consistent readings. Despite its small range, alpha radiation is deadly if ingested or inhaled making accurate surveys critical for worker safety. The drudgerous and inefficient routine surveys are typically performed by dedicated Radiation Contamination Technicians (RCTs). Mobile robotics provides a promising alternative to this process, capable of providing services which augment and accelerate the process alongside human workers. 

Mobile systems have previously been considered as strong candidates for radiation monitoring. Marques et al. \cite{MDPI_Paper} published a comprehensive survey of different manned and unmanned mobile systems designed for gamma and neutron radiation survey. However, systems designed for alpha monitoring were excluded given the difficulty introduced by the limited detection range. A research group from the University of Manchester has developed a system capable of monitoring alpha, beta, and gamma radiation called the CARMA II system \cite{NewAlphaPaper}. This system was designed to collect data on radiation levels while preventing the spread contamination caused by driving through contaminated areas. However, this system can only detect alpha radiation at floor level and requires human input to assist in path planning.

Designing a single robot, which is capable of performing a full radiation survey is not practical. Different aspects of surveys benefit from different strategies, each of which requires unique software and hardware. As such, the survey is best distributed across a team of custom mobile robots. For example, completely surveying a facility floor for alpha contamination can take hours to complete at the required uncertainty level even with a large sensor, while elevated surfaces within the workspace can be examined much faster but require a smaller sensor mounted on a dexterous mobile manipulator. The former task is optimally executed by a wheeled mobile robot due to the exceptional energy efficiency and steady movement characteristics of wheeled bases, while the latter can be assigned to a quadrupedal mobile manipulator which can take advantage of its large number of degrees of freedom, despite having a much shorter battery life. Small robots would enable surveying difficult or impossible-to-reach areas. Robotic surveys also enable detailed archiving of exact survey locations for reporting and temporal comparison. Other tasks such as robot collaboration, gamma/beta localization, waypoint planning, mapping, and exploration all require custom solutions working in tandem to execute a single autonomous survey.

This paper presents an overview of an ongoing project to provide a ``full stack'' hardware and software survey solution, developing a team of heterogeneous robots that collectively provide a complete and autonomous survey of an active nuclear production or research facility. The high-level requirements include:
\begin{itemize}
    \item Generate heat maps in lab space for beta/gamma radiation in a repeatable fashion.
    \item Generate heat maps for alpha radiation in any location that could be surveyed by an RCT. This would include high-traffic areas, common-contact areas, and locations such as under desks or along floor boards typically reached with a swipe connected to a long handle. The system ideally reaches locations the RCTs cannot reach.
    \item While surveys ideally happen at night, the navigation systems must be robust to dynamic objects including humans.
    \item The system must be able to clear the space of static obstacles such as chairs or trash cans.
    \item The coverage maps must be comprehensive and repeatable.
    \item The robot team must be able to complete full surveys on a routine (nightly) basis with no human intervention necessary. 
\end{itemize}

The system overview outlined below documents efforts completed at the University of Texas at Austin \cite{anderson_optimization_2022-1}, \cite{anderson_mobile_2020}, \cite{anderson_generating_2017}, \cite{pitsch_self-navigating_2018} and elsewhere as comprehensively documented by \cite{MDPI_Paper} and recent efforts by \cite{Tsitsimpelis2019-hw}, \cite{NewAlphaPaper} and others demonstrate that radiation surveys are possible. It is the goal of this effort to present our current results as well as our roadmap for meeting the high-level requirements outlined above using a heterogeneous team of robots in a nuclear facility. The following sections outline key technical efforts developed for the integration and execution of the storyboard outlined in Section \ref{sec:integrated_demo_description}. We then conclude with a quick discussion on the next steps and remaining technical issues that must be addressed.

\section{System Overview}
\label{sec:system_overview}

Three heterogeneous robotic systems are utilized to complete an alpha/beta/gamma radiation survey in an active nuclear facility. Robots are referred to as robo-RCTs and collectively as the Robot Radiation Survey System (RRSS)
\begin{itemize}
    \item \textbf{Minibot} - Low clearance, low-cost robots for surveying and mapping in difficult-to-reach places.
    \item \textbf{Magni} - A wheeled platform with a large alpha sensor to efficiently survey high-traffic areas.
    \item \textbf{Alph} - A quadruped system to survey elevated and common-contact surfaces. Using UT's developed screw-based affordance templates \cite{pettinger2022}, Alph can clear obstacles allowing Minibot and Magni to complete their surveys.
\end{itemize}

The robotic hardware and selected enabling algorithms are discussed in the remainder of this section. The systems will utilize beta/gamma survey, alarm, and localization algorithms previously developed by UT Austin. \cite{anderson_optimization_2022-1} \cite{2018CoverageSurvey} \cite{anderson_mobile_2020} \cite{anderson_generating_2017}

\begin{figure}[b!] 
    \centering 
    \includegraphics[height=4.5cm, frame]{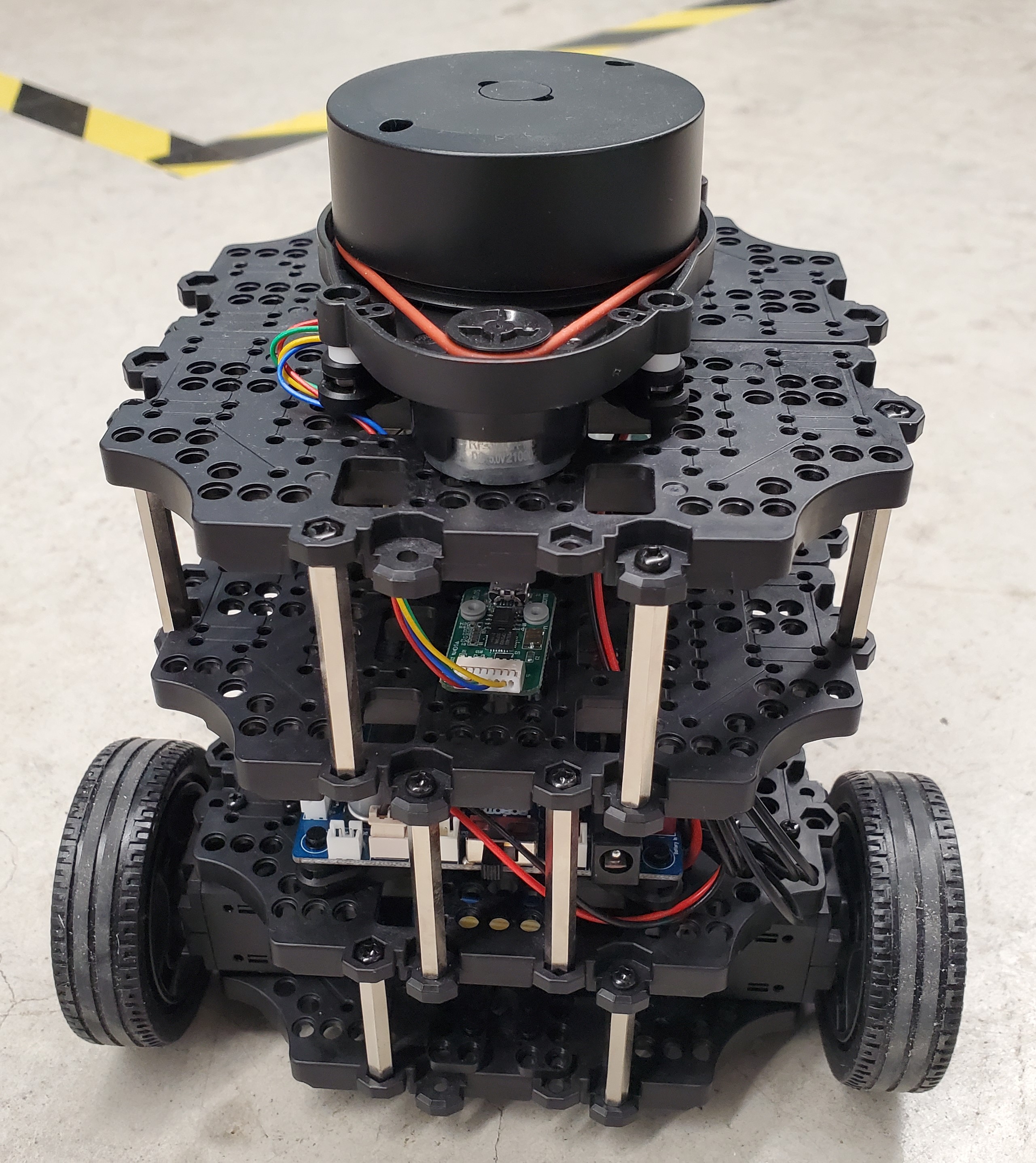}
    \caption{Minibot} 
    \label{fig:turtlebot_alone_close} 
\end{figure} 

\renewcommand{\paragraph}[1]{\smallskip\noindent\textbf{#1}}
\paragraph{Minibot(s):} The Minibot System is comprised of one or multiple small robot agent(s) that work in collaboration with a remote server computer to traverse designated environments and generate detailed 3D maps of the nuclear facilities inclusive of areas below desks and lab equipment. These maps can be integrated with other RRSS collected map data. We are currently using a TurtleBot3 Model Burger [Figure \ref{fig:turtlebot_alone_close}] that will be modified to meet dimension and sensor requirements.

\begin{figure}[!h]
    \centering
    \includegraphics[height=4.5cm, trim={2cm, 1cm, 3cm, 2cm}, clip]
    {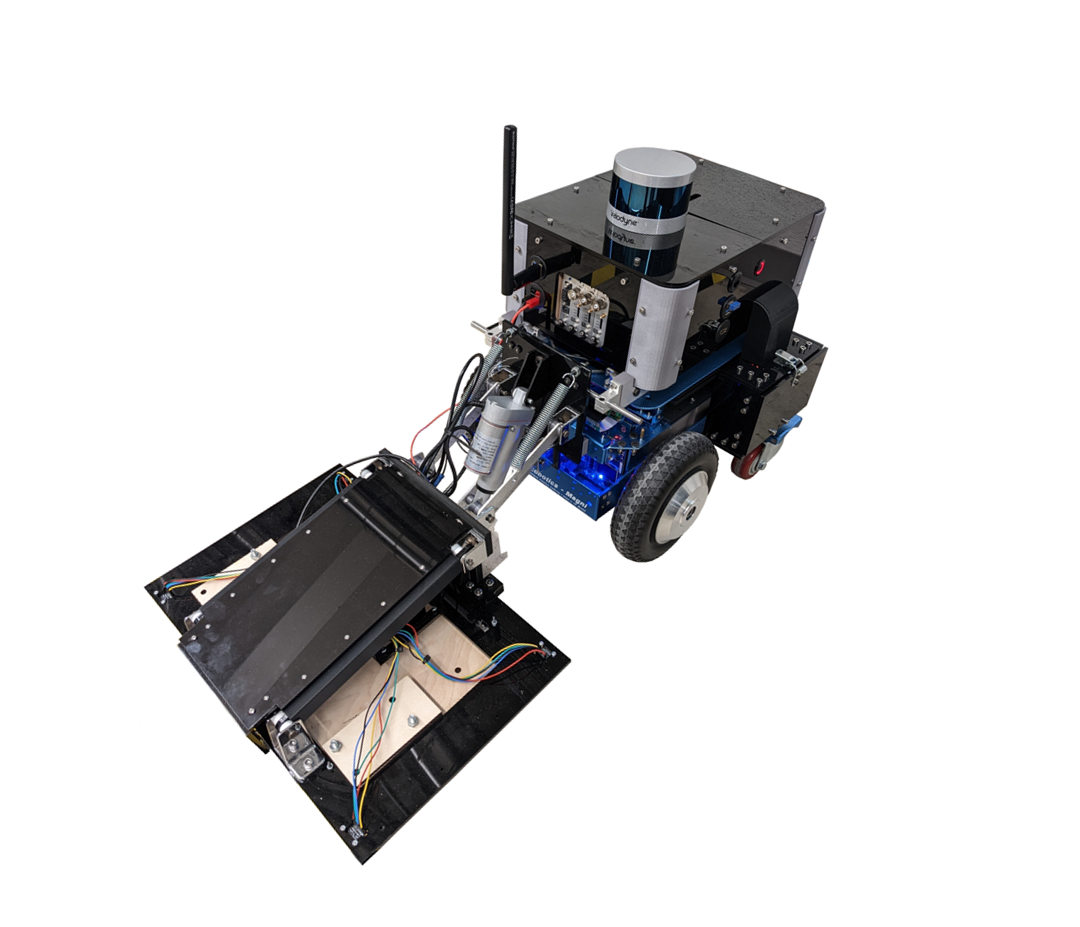}
    \caption{Magni Robot}
    \label{fig:magnibot}
\end{figure}

\paragraph{Magni:} A differential drive mobile robot by Ubiquity Robotics, Magni [Figure \ref{fig:magnibot}] is customized with a trailer module that contains an air-proportional detector for detecting alpha-emitting particles. Magni's main role is to autonomously survey the floor for contamination. It does so by employing a coverage algorithm that ensures the detector has covered all available floor space. Additionally, Magni will avoid areas it hasn't surveyed to prevent the spread alpha-emitting particles. Once the survey is completed, it provides technicians with a heat map for comparison to previous surveys.   

\begin{figure}[h!] 
    \centering 
    \includegraphics[height=4.5cm, frame]{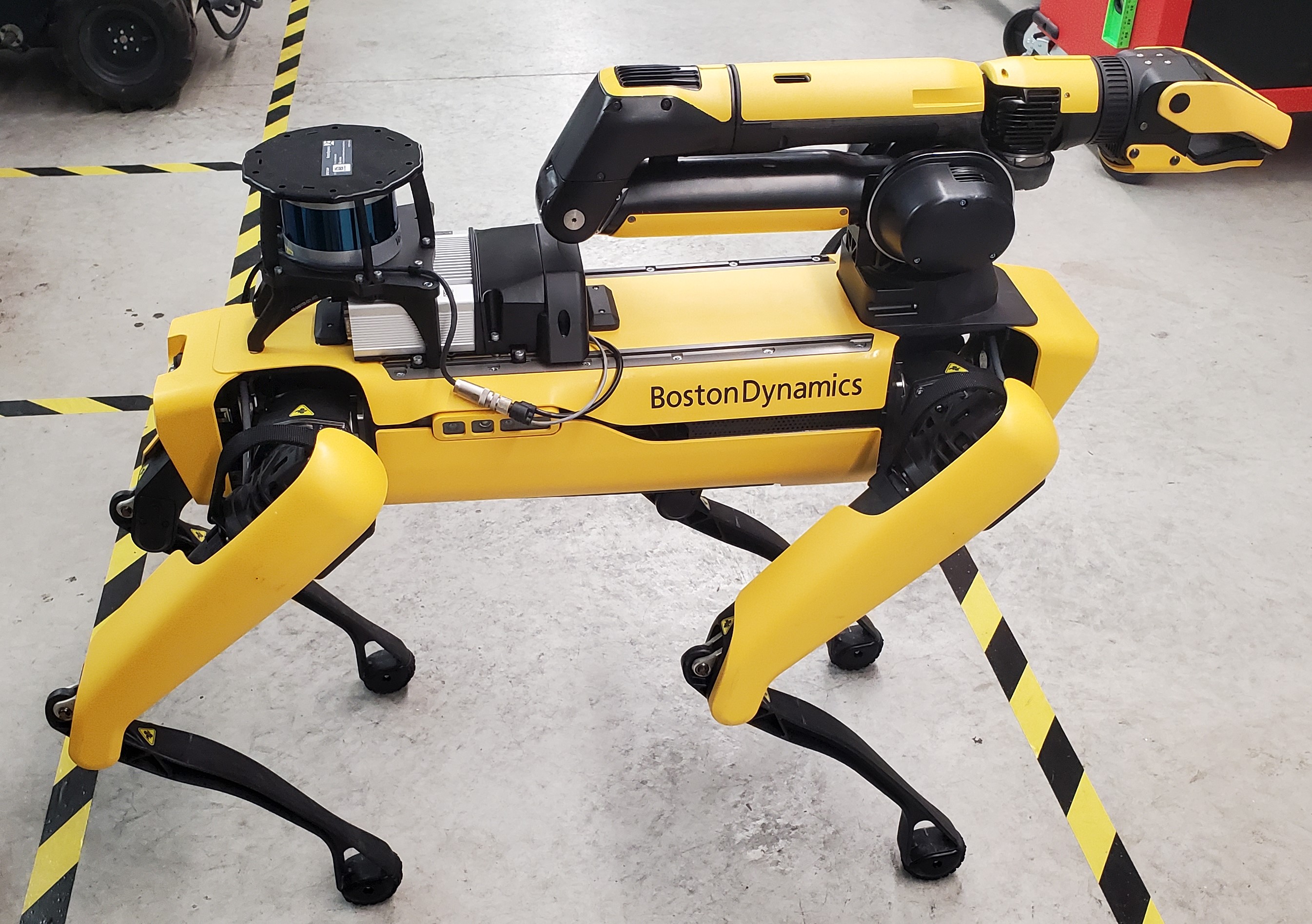} 
    \caption{\centering Manipulator-integrated Alph} 
    \label{fig:Spot} 
\end{figure}

\paragraph{Alph:} Another robot is needed to ensure that all reachable surfaces are included in the survey and to remove static obstacles from the paths of the wheeled platforms. Alph is a manipulator-integrated Boston Dynamics Spot quadruped robot [Figure \ref{fig:Spot}] capable of traversing uneven terrains and manipulating objects with a six DOF arm and gripper \cite{boston_dynamics_arm_specs}. Alph performs radiation-survey-related tasks including surveying elevated surfaces, opening doors, and moving static obstacles.

\section{System Components And Results}
\label{sec:system_components_and_results}

This section cannot include all the algorithms and capabilities necessary to deploy the RRSS, but new or unique algorithms developed and necessary for a ``full-stack'' solution are review in this section.

\paragraph{Multi-Robot Mapping and Exploration:} The Minibots are the first to traverse unmapped and unexplored areas, that may present nuclear contamination, to create necessary 2D and 3D maps of the environment. Thus, it is possible that a robot agent may get contaminated or physically damaged during its operation. This motivates the use of expendable platforms that can be easily repaired, retrieved, or discarded when an incident occurs. These systems however have limited computer hardware that does not have the capability of generating point cloud maps. To alleviate this problem, we disassociate the physical task execution from the processing operations of the system by having the robot agent perform the physical actions while the server computer executes all necessary calculations remotely.

The robot agent(s) is currently built on a Turtlebot3 Model Burger platform \cite{Turtlebot3Burger}, though smaller hardware systems are in consideration. The robot is small enough to reach places not accessible by Magni and Alph, including underneath tables, behind furniture, and through small gaps; which enables it to generate more complete representations of the environment including 3D maps of difficult-to-reach locations. These are then accessed by any RRSS system sensors to survey areas reachable by their detectors even if inaccessible to the full robot platform. For example, the Magni's wide front trailer can be inserted under a lab desk as shown in \ref{sec:integrated_demo_description}.

The server computer is in a remote location under the same Local Area Network as the robot agent. It communicates wirelessly with the agent to send movement commands and receive relevant sensor information, which it uses to run computationally complex localization and mapping algorithms. Assuming the network is secure, overall information security is improved since no MiniBot maintains historical information about their surroundings, rather they simply transmit current readings and allow the server to handle all incoming data. Additionally, due to their low perspective, commonly used work areas, such as tables and desk tops, are not included in the generated maps.

\paragraph{Planar Coverage Path Planning: } The forward-mounted location of the alpha detector on the Magni robot provides advantages in safety and reachability by placing the sensor away from the robot body. To fully take advantage of this sensor geometry, 3D obstacle detection is required to allow the sensor to be inserted under overhanging obstacles such as tables and chairs, as well as to allow it to pass over low-lying obstacles like wire guards. However, accounting for 3D obstacles with this sensor geometry complicates the planning process. 

Coverage planning algorithms generate high-level motion plans for mobile robots which ensure that as much of the workspace as possible is visited by the robot. Existing coverage algorithms typically generate plans for the robot center and do not provide considerations for off-center tools \cite{2013CoverageSurvey}. Some works have proposed solutions to this problem such as planning for the sensor and later solving for reachable base-link poses \cite{2018CoverageSurvey}, but these solutions work best for 2D obstacle representations \cite{NavarroCoverage}. 3D coverage path planning generally applies to manipulators covering a 3D surface or drones scanning 3D objects \cite{2013CoverageSurvey}, and so is not applicable to the case of planar coverage planning while accounting for 3D obstacle geometry. Therefore a new coverage planning framework was needed which accounts for both the off-center radiation detector and a 3D representation of the environment.

\begin{figure}[ht]
    \centering
    \includegraphics[height=4.5cm, frame]{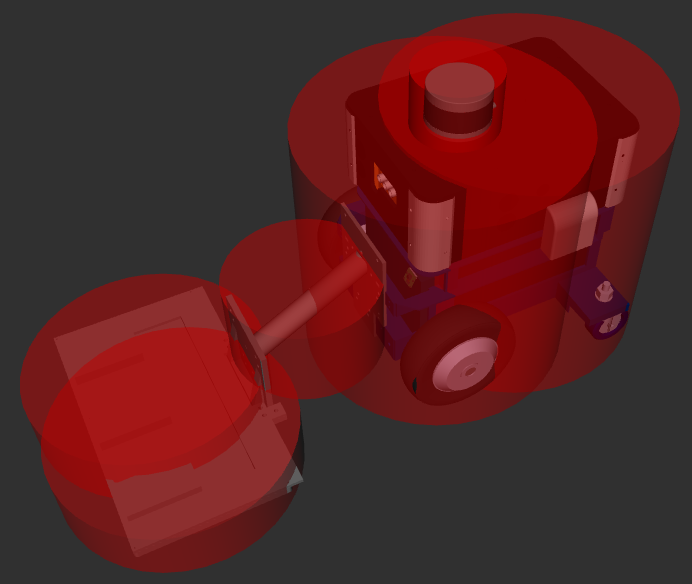}
    \caption{Cylinder Decomposition of Magni Robot}
    \label{fig:CylinderDecomposition}
\end{figure}

The issue of 3D obstacle detection was solved with the use of \textit{cylindrical robot decomposition}. The robot is represented by a set of cylinders which collectively cover the robot geometry, as seen in Figure \ref{fig:CylinderDecomposition}. Each of these cylinders is considered as a separate collision entity, which can each be tested against an input pointcloud to determine if the robot state is valid. This allows the sensor geometry to occupy space which would be considered a collision state for the robot body, and vice versa.

The off-center tool was accounted for through the use of \textit{grid cell partitions}. Consider a regular 2D grid overlaid on the workspace. Instead of populating the grid entries with obstacle values, as in common in 2D navigation, we instead further subdivide each grid cell into the set of allowable orientation ranges $\mathbb{R} = \{R_1, R_2, ..., R_N\}$ for the robot at that cell. Each range $R_i = [\phi_{i1}, \phi_{i2}]$ is defined such that if the robot sensor is placed over the grid cell in question and is oriented at an angle $\theta$ such that $\phi_{i1} \le \theta \le \phi_{i2}$, then the robot is guaranteed to be in a valid state. Each such range is labelled a partition $P(x, y, i)$ representing the $i^{th}$ range $R_i$ for the cell at index $(x, y)$.

\begin{figure}[b!]
    \centering
    \includegraphics[width=\linewidth]{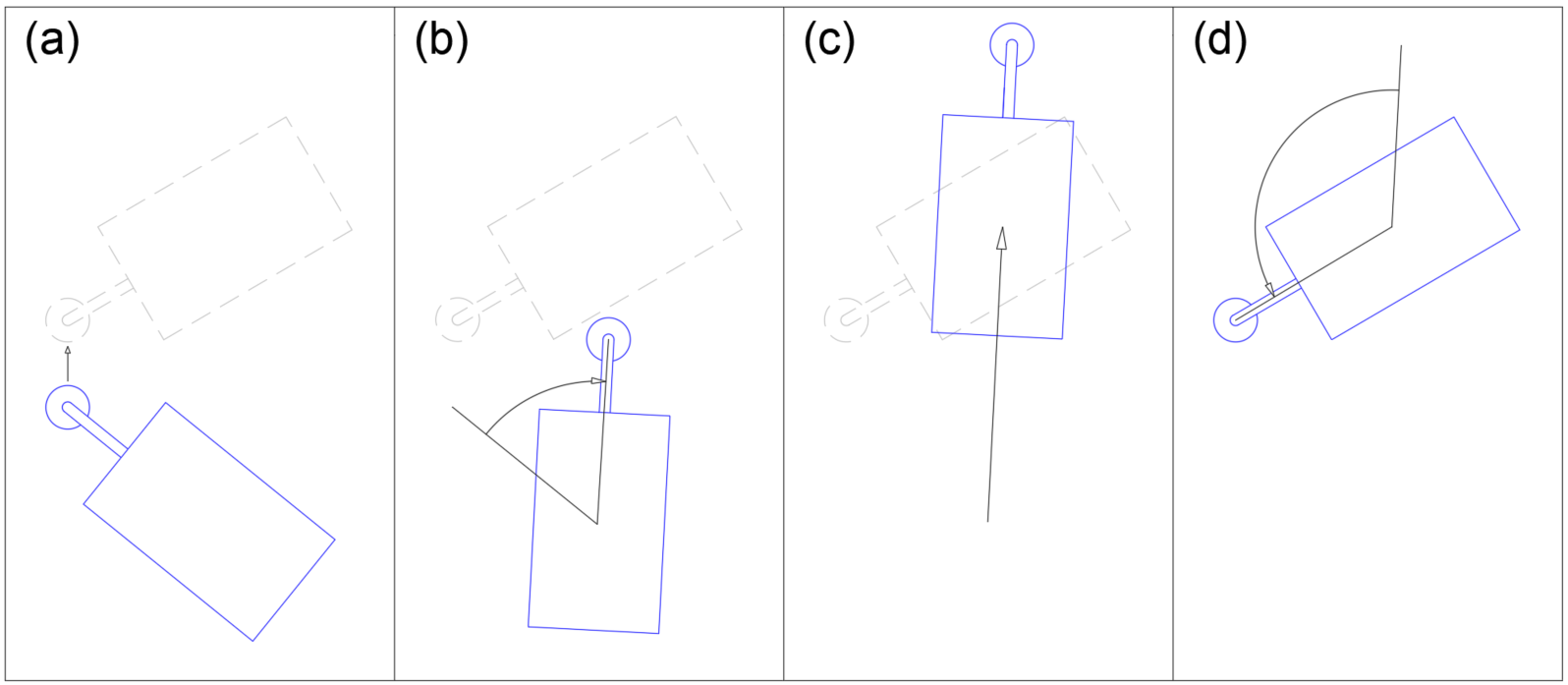}
    \caption{Kinematic Transition Test}
    \label{fig:transition}
\end{figure}

Each of these partitions is now inserted as a node into a navigation graph $\mathbb{G}$. Edges are inserted between the nodes of $\mathbb{G}$ if the corresponding partitions are grid-wise adjacent and if it is kinematically possible for the robot to transition from the first partition to the second. This kinematic feasibility check is executed through simple forward simulation of the robot turning to face the final base-link location, approaching on a straight line path, and finally turning to the final orientation as in Figure \ref{fig:transition}.

As a result of this process, the graph $\mathbb{G}$ represents a fully defined set of valid robot states and ways to transition between them. $\mathbb{G}$ can be substituted in place of a traditional 2D grid into most coverage planning algorithms such as the Path Transform algorithm \cite{GridBasedCoverage} or the Backtracking Spiral algorithm \cite{BSA_Gonzales2003} to obtain sensor-aligned coverage plans which account for the kinematic constraints of the robot and its full 3D geometry.

\paragraph{Robot Velocity Optimization: } Current alpha contamination detection methods involve swiping a surface and moving the swipe across a radiation detector at 2 in/s. If the detector reads above a given count threshold, then the \textit{entire} area swiped is considered contaminated. 
Through the use of a larger sensor like the one in Figure \ref{fig:magnibot}, Magni could achieve readings within the desired uncertainty range at a faster pace. The velocity at which the detector, and thus Magni, can move is quantifiable, repeatable, and depends on a variety of factors such as source strength, detector area, detector efficiency, and detection policy limits. The velocity can be optimized such that confidence in detection is known and high. When a count rate measurement meets the desired precision requirement in counts per second, it is then compared to the count threshold, 
\begin{equation}\label{ct}
 CT = L_{D}A_{d}\epsilon_{d}
\end{equation}
where $L_{D}$ (dps/cm$^2$) is the facility's detection policy, $A_{d}$ (cm$^2$) detector area, and $\epsilon_{d}$ is detector efficiency. If the measurement is above the count threshold, then the area that produced that count rate will be deemed contaminated.     

\begin{figure}[!h] 
    \centering 
    \includegraphics[width=1\linewidth]{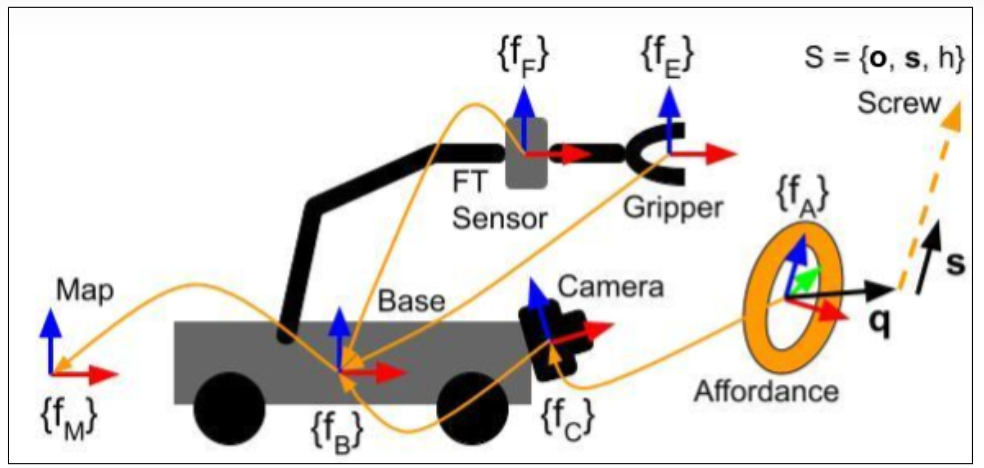} 
    \caption{\centering Illustration of Pettinger et. al.'s AT framework using screw theory to rotate a wheel valve with a manipulator. The framework uses a screw axis $\mathbf{s}$ passing through a point $o$  and including the pitch $h$ that represents the ratio of translation to rotation. The vectors $\mathbf{s}$ and $\mathbf{o}$ are defined with respect to the frame ${\{f_A\}}$ \cite{PettingerPhDThesis}. }
    \label{fig:affordance} 
\end{figure}

\paragraph{Affordance Templates:} Although a BD Spot comes with useful commercial capabilities such as opening doors and turning valves, they are computationally expensive, limited, and cannot be extended to new tasks such as autonomously moving a chair out of the way \cite{boston_dynamics_constrained_manipulation}. Therefore, a computationally efficient and adaptable framework is needed to perform a variety of contact tasks. Affordance Templates offer such a framework.
The term "affordance" was coined by psychologist J. J. Gibson to describe structures that offer functional merit to the environment, such as a book affording reading or a hammer affording hitting \cite{Gibson}. In robotics, affordance templates (ATs) are 3D representations of objects and their interaction methods. Initially developed by a NASA-JSC DARPA Robotics Challenge team, ATs have since been successfully implemented in various platforms. However, their original form involves describing object interactions using numerous waypoints \cite{Hart}, which is computationally inefficient. To remedy this, Pettinger et. al. replaced the waypoints with a Screw framework that only requires four parameters to describe it in an object-centric way [Figure \ref{fig:affordance}]. We implement this version of ATs on Alph to carry out the contact tasks necessary for radiation surveys.

\paragraph{Precision Manipulator Control for Contact Radiation Detection:} In addition to clearing a path for the Magni system, Alph has the additional job of measuring radiation levels on surfaces such as tabletops or the exteriors of glove boxes. To do this, Alph uses radiation swipes to wipe surfaces and then holds these swipes over a detector before disposal in a nuclear waste bin. Each swipe is a 4.5 cm diameter cloth disk specifically designed for radiation collection and is designed to be used for testing approximately one square meter of surface. These swipes are fragile, so the force exerted between the robot and the surface it is detecting must be carefully controlled as to wipe with enough force to lift radioactive dust from the surface, but not enough force that the swipe is damaged. The applied force is determined using the force-torque sensor integrated into the Spot Arm end effector and is controlled using a PID controller and setting the maximum applied force using the Boston Dynamics API.

\paragraph{Task Virtual Fixtures: } Virtual Fixtures (VFs) are constraints on a robot’s operation useful for semi-autonomous tasks represented by static sensor information fed to the robot that overlays the sensor feedback. This can be thought of as a virtual ruler that would allow a human to draw a straight line \cite{Rosenberg}. Two of the key types of VFs are Guidance Virtual Fixtures (GVFs) and Forbidden Region Virtual Fixtures (FRVFs). GVFs use virtual surfaces to help guide the movement and operation of robots by excluding certain regions known (FRVFs) \cite{Sharp}. This allows the human operator to designate a task and the robot to more accurately perform the task using GVFs to constrain its motion and performance to the general desired motion.

When it comes to tasks that require surface coverage, one of the more complicated variables is the distance between the end effector and the surface. VFs can be used to eliminate this variable entirely by using a desired normal distance from the surface to constrain movement to only the surface at that offset. This is achieved by creating several VFs at set offsets from the surface and designating some layers to be FRVFs and one or more to be GVFs. This combination of polygonal mesh FRVFs and point cloud layered GVFs is known as a Task Virtual Fixtures (TVF) \cite{Sharp}.

\begin{figure}[ht]
    \centering
    \includegraphics[width=.6\linewidth]{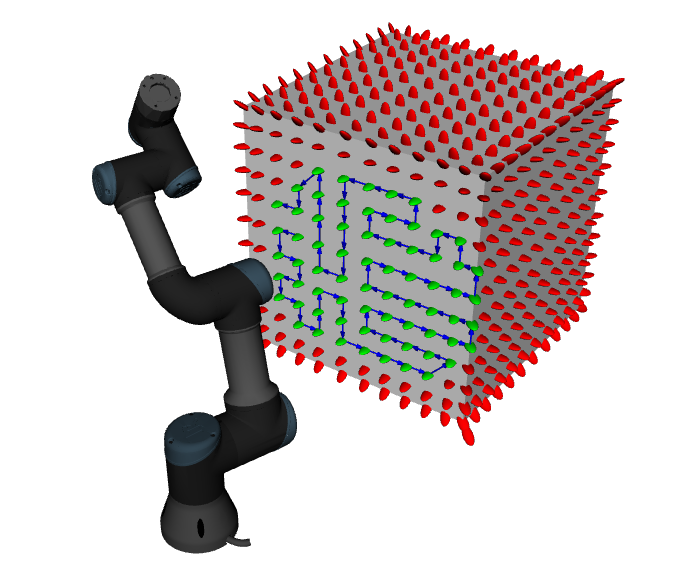}
    \caption{Simulated TVF on cube surface with Universal Robotics UR5 Robot}
    \label{fig:TVF_example}
\end{figure}

TVFs also contain the ability to path plan for coverage of the desired surface using the traveling salesperson problem methodology to generate a path between the TVF point cloud points. For our intended survey applications, the ability to plan a path that guarantees full surface coverage is paramount. Using TVFs to generate this path that also maintains a constant distance from the surface is important when performing a task such as radiation detection to ensure uniform detection results.

\begin{figure}[!b] 
    \centering 
    \includegraphics[width=0.83\columnwidth, frame]{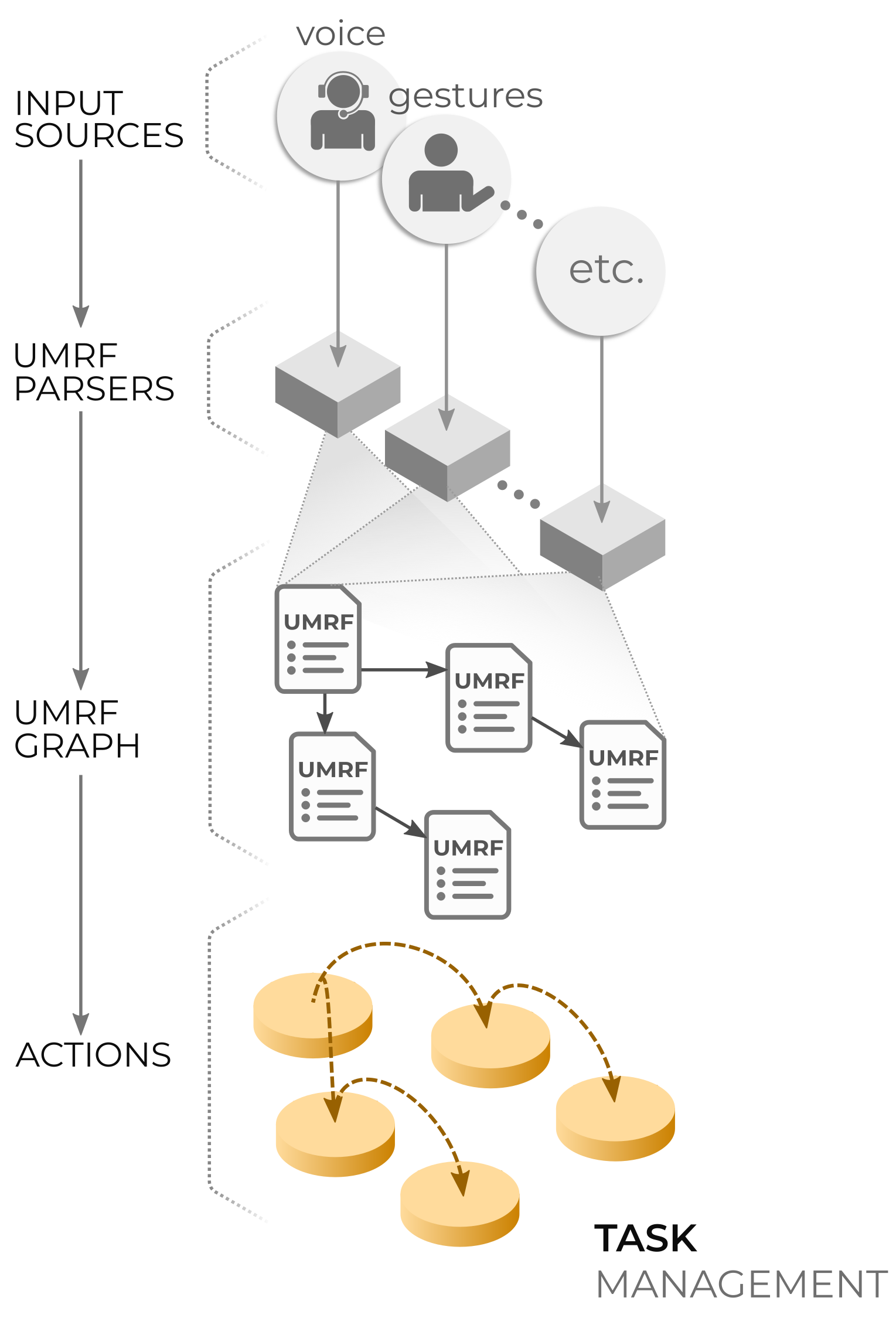}
    \caption{TeMoto for fault-tolerance and dynamic reconfiguration \cite{valner_temoto_2021} } 
    \label{fig:temoto_graphic} 
\end{figure} 

\paragraph{Temoto:} Fault tolerance is essential for safe operation given the chance for radiation exposure, the
long and routine survey durations, and the inevitable need to adapt to new tasks. While the robotic system is not designed for emergency response, it would be unfortunate if members of the RRSS could not be used remotely (even as tele-operated systems) to mitigate any contamination event prior to personnel entering the area. Fault tolerance and adaptive behaviors have been well-investigated on a component level but receive little attention at the system level in robotics. To address this, we utilize TeMoto \cite{valner_temoto_2021}, a novel architecture for adaptive autonomous robots, and a ROS-based framework of openly available software tools that implement the TeMoto architecture. The TeMoto enables run-time reconfiguration of one or more robots for common scenarios (teleoperation, autonomous surveillance, cargo delivery, etc.). TeMoto-based systems exhibit fault tolerance and \textit{dynamic} reconfigurability of software and hardware resources. 

\section{Integrated Demo Description}
\label{sec:integrated_demo_description}

\begin{figure}[!h] 
    \centering 
    \includegraphics[width=.8\linewidth, frame]{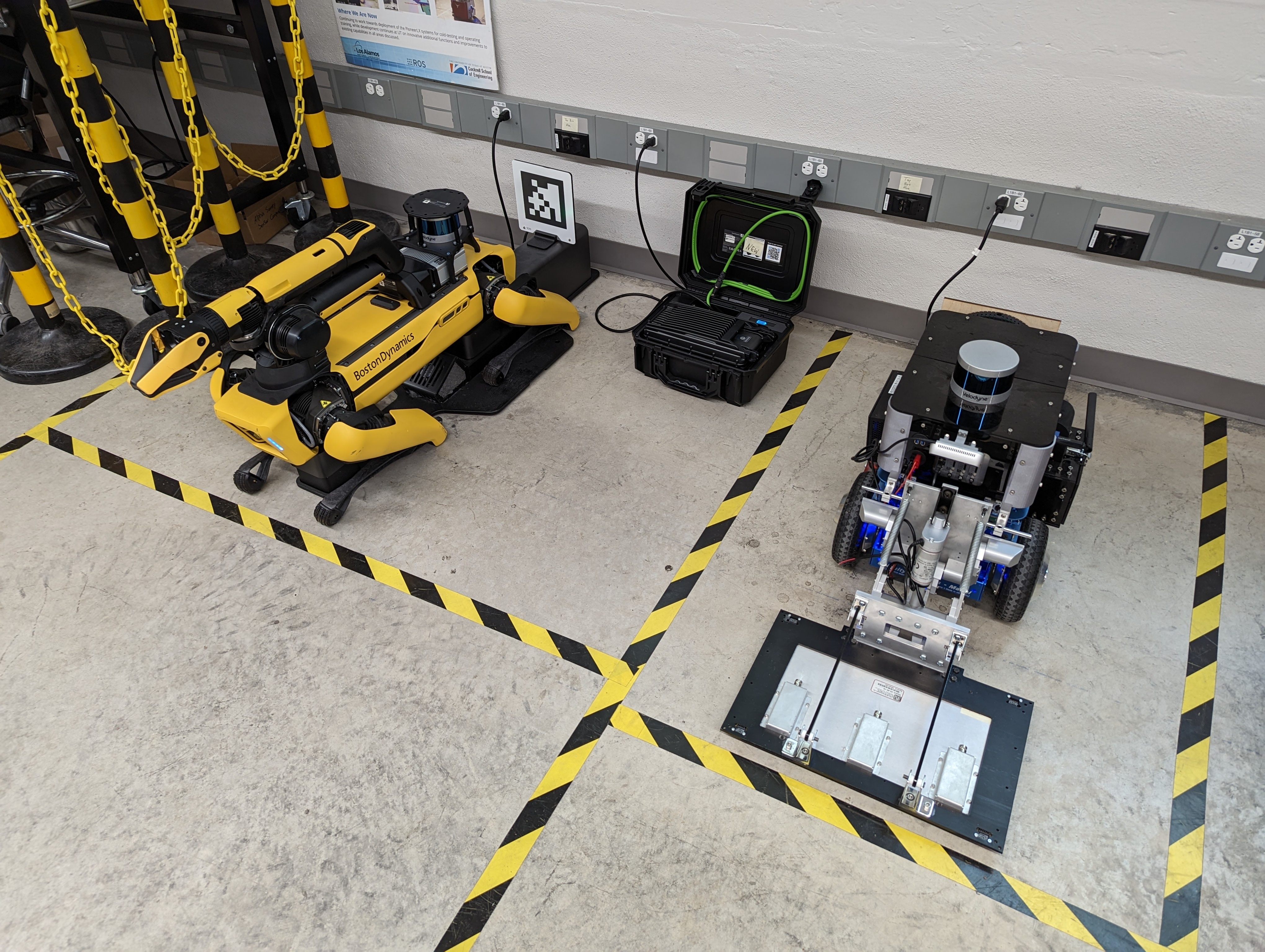} 
    \caption{\centering Docked Robots} 
    \label{fig:docking_station} 
\end{figure}

\begin{figure}[!b] 
    \begin{subfigure}[t]{.49\linewidth}
    \centering 
    \includegraphics[width=1\linewidth, trim={20cm, 15cm, 0, 0}, clip, frame]{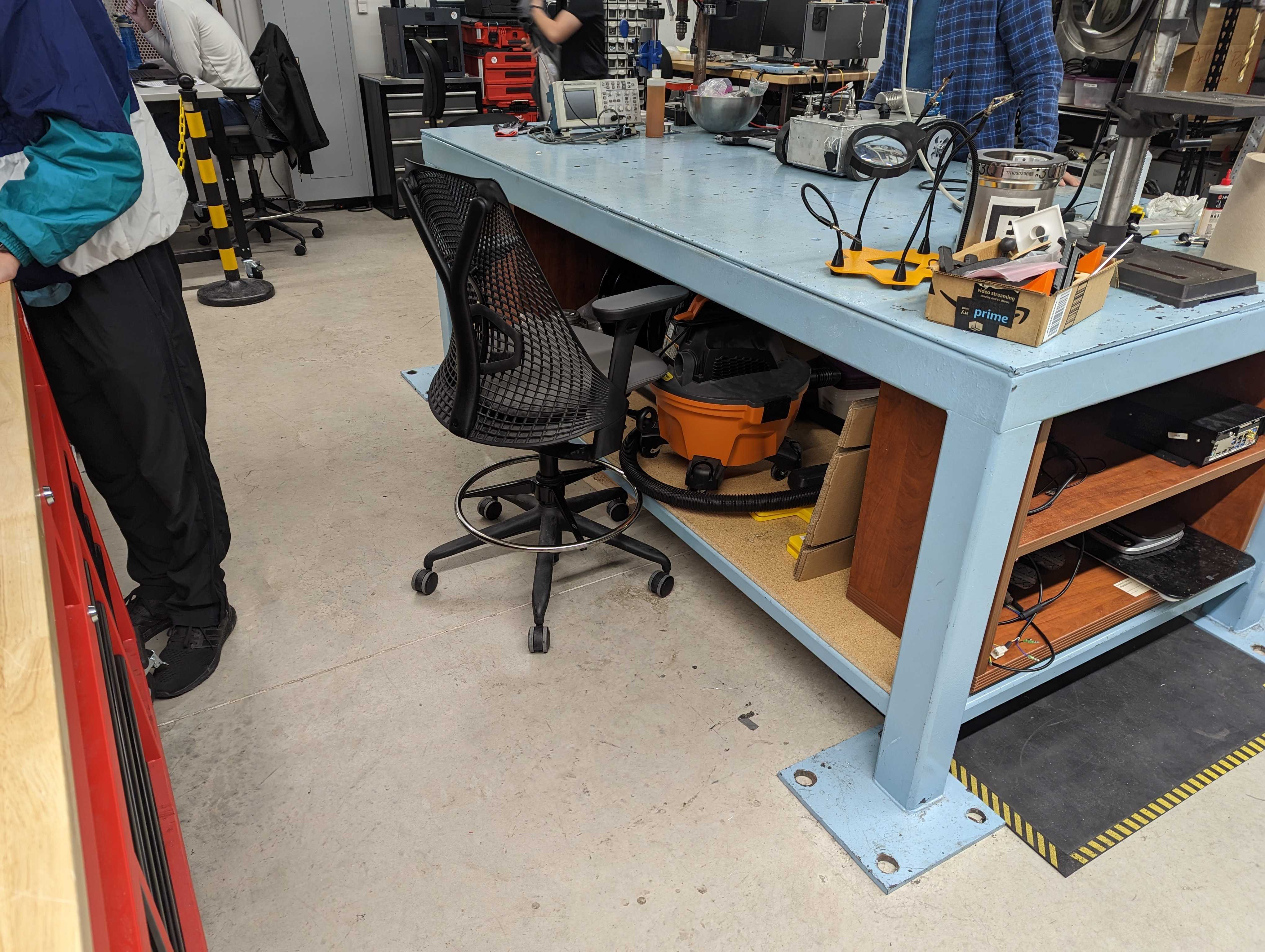}
    \caption{\centering Movable Obstacle Detected} 
    \label{fig:spot_moving_chair:a}
    \end{subfigure}
    \hfill
    \begin{subfigure}[t]{.49\linewidth}
    \centering 
    \includegraphics[width=1\linewidth, frame]{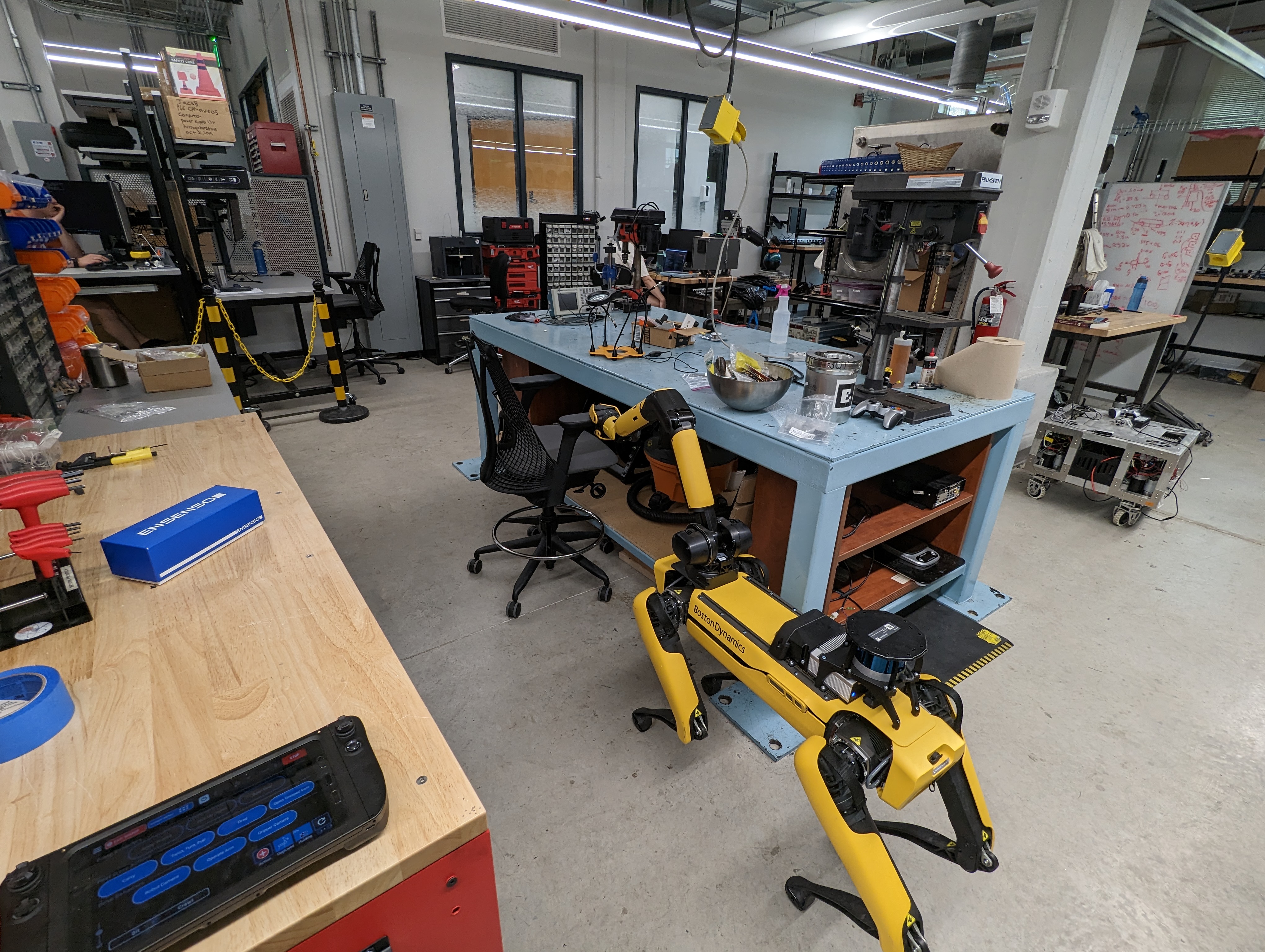}
    \caption{\centering Alph Deploys Manipulator} 
    \label{fig:spot_moving_chair:b}
    \end{subfigure}

    \begin{subfigure}[t]{.49\linewidth}
    \centering 
    \includegraphics[width=1\linewidth, frame]{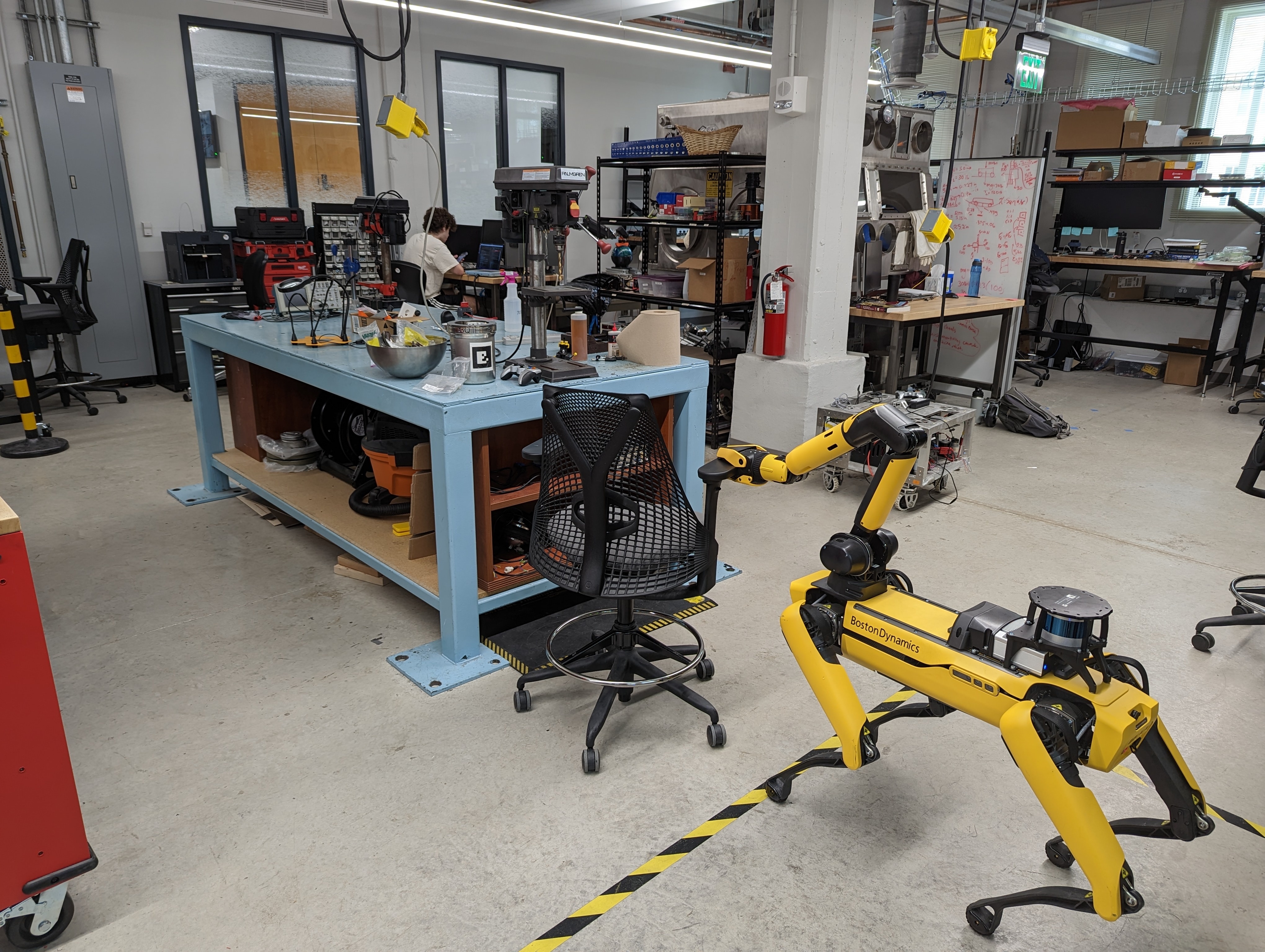}
    \caption{\centering Obstacle is Removed} 
    \label{fig:spot_moving_chair:c}
    \end{subfigure}
    \hfill
    \begin{subfigure}[t]{.49\linewidth}
    \centering 
    \includegraphics[width=1\linewidth, frame]{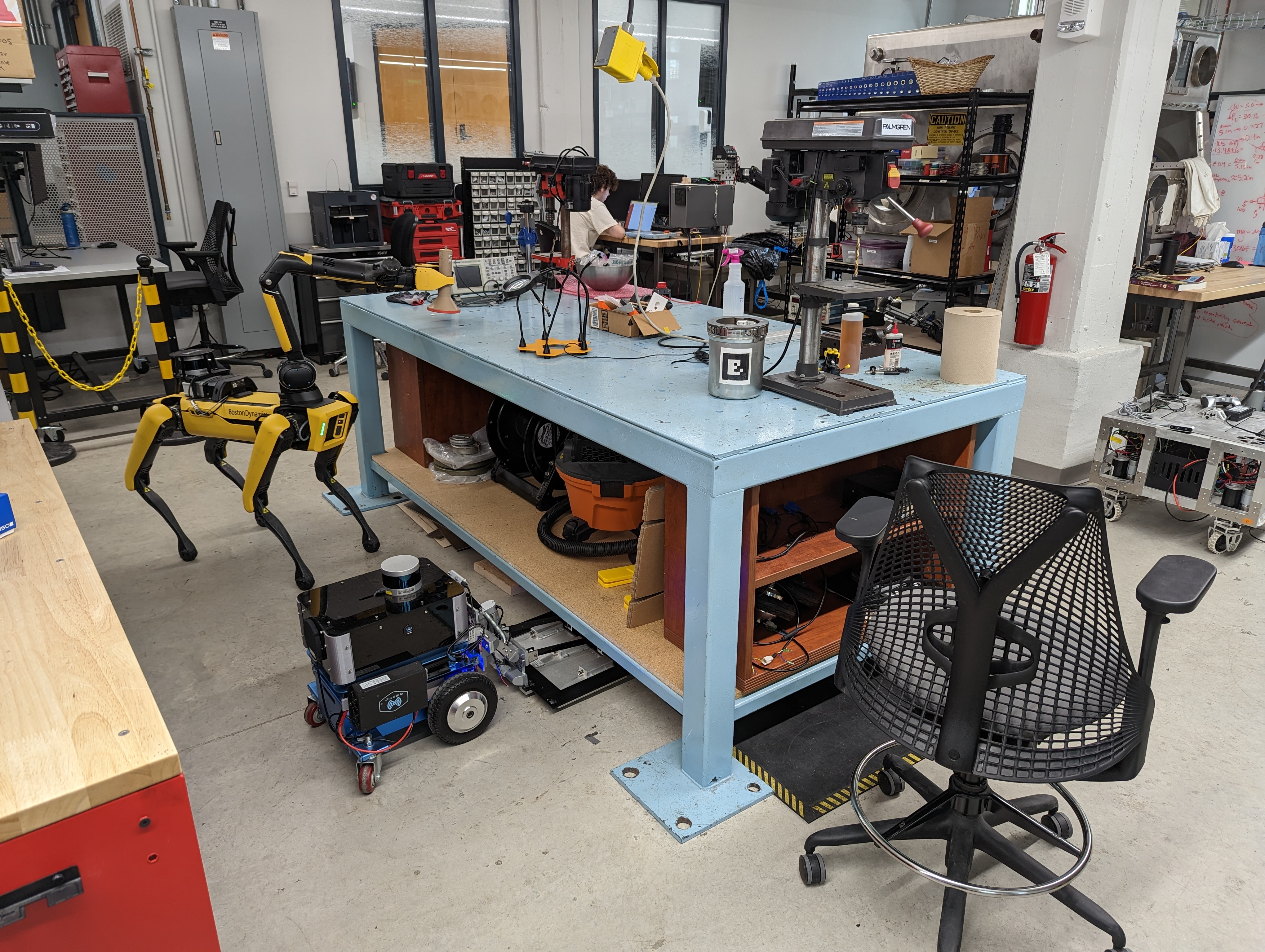}
    \caption{\centering Both Robots Resume Survey} 
    \label{fig:spot_moving_chair:d}
    \end{subfigure}
    
    \caption{\centering Heterogenous collaboration allows Magni to request help moving lightweight obstacles from its path} 
    \label{fig:spot_moving_chair} 
\end{figure}

\begin{figure}[!b] 
    \centering 
    \includegraphics[width=.8\linewidth, frame]{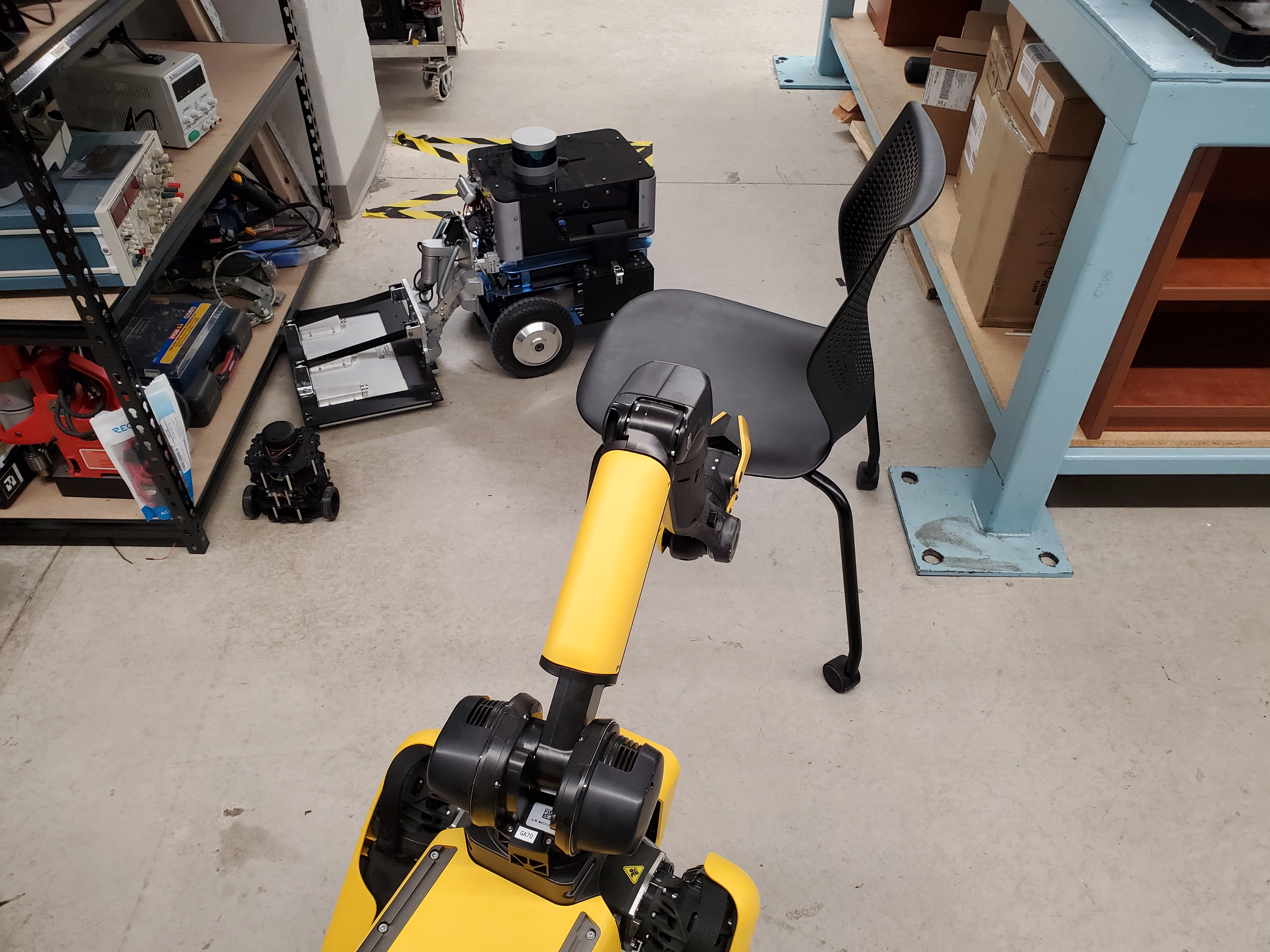} 
    \caption{\centering Simultaneous Survey Execution and Collaboration in an Integrated Demo} 
    \label{fig:all_three_robots} 
\end{figure}

Figures \ref{fig:docking_station} thru \ref{fig:all_three_robots} summarize key elements of an integrated survey using the RRSS integrated team of heterogeneous robots. These images are generated from existing capabilities such as the robots' autonomous docking and undocking [Figure \ref{fig:docking_station}], and the utilization of an affordance of a chair to move it to the side [Figure \ref{fig:spot_moving_chair:c}]. Others [Figure \ref{fig:all_three_robots}] were created to storyboard the full demo and better capture the full system requirements. For example, we are currently able to move a chair or a trash can, but a high-level open question remains: where should these objects be moved to?  

\begin{figure}[!b] 
    \centering

    \begin{subfigure}[t]{.49\linewidth}
        \centering 
        \includegraphics[width=1\linewidth, frame]{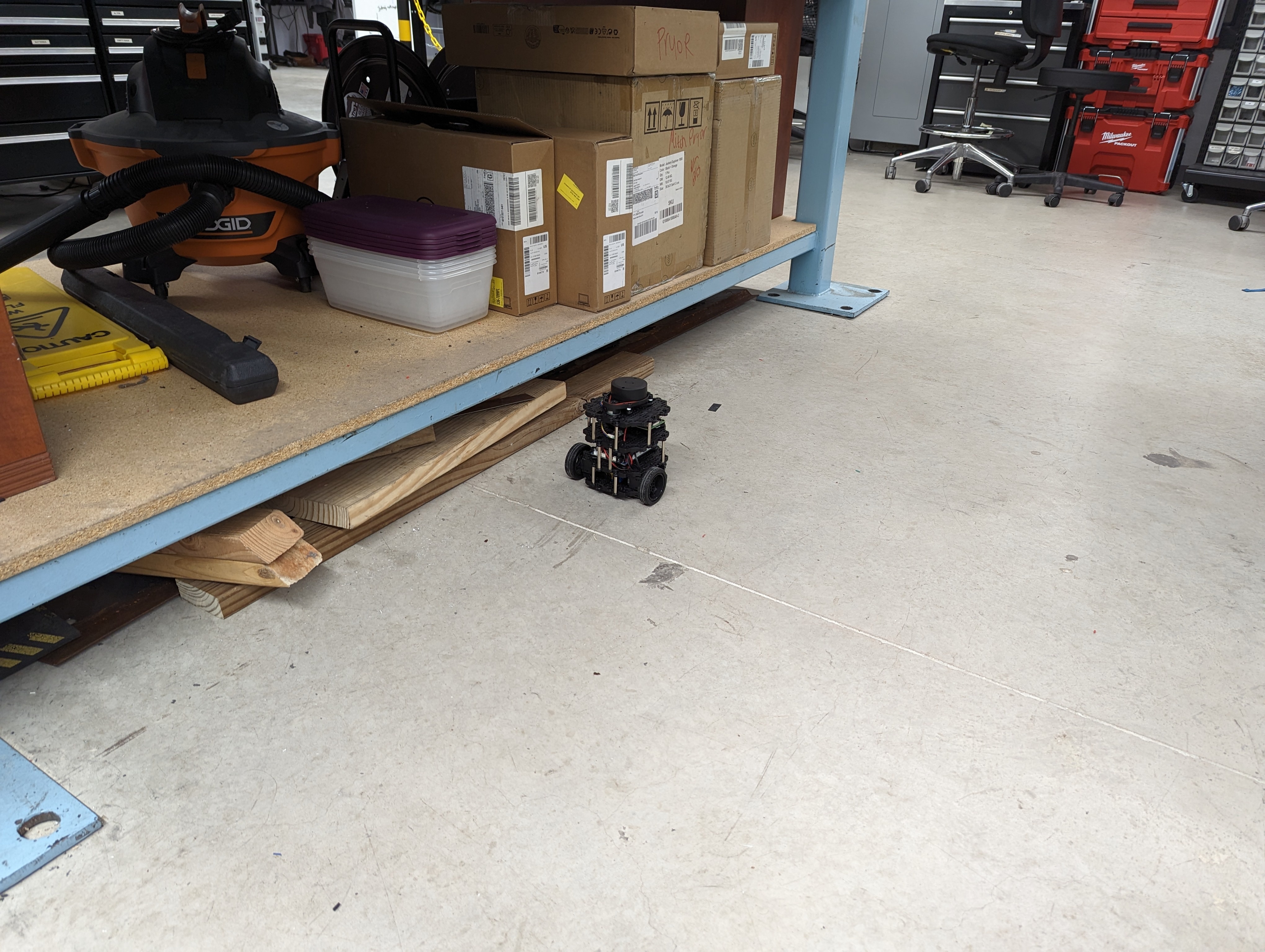}
        \caption{\centering Low Obstacle Detection} 
        \label{fig:minibot_small_obstacle:a} 
    \end{subfigure}
    \hfill
    \begin{subfigure}[t]{.49\linewidth}
        \centering 
        \includegraphics[width=1\linewidth, frame]{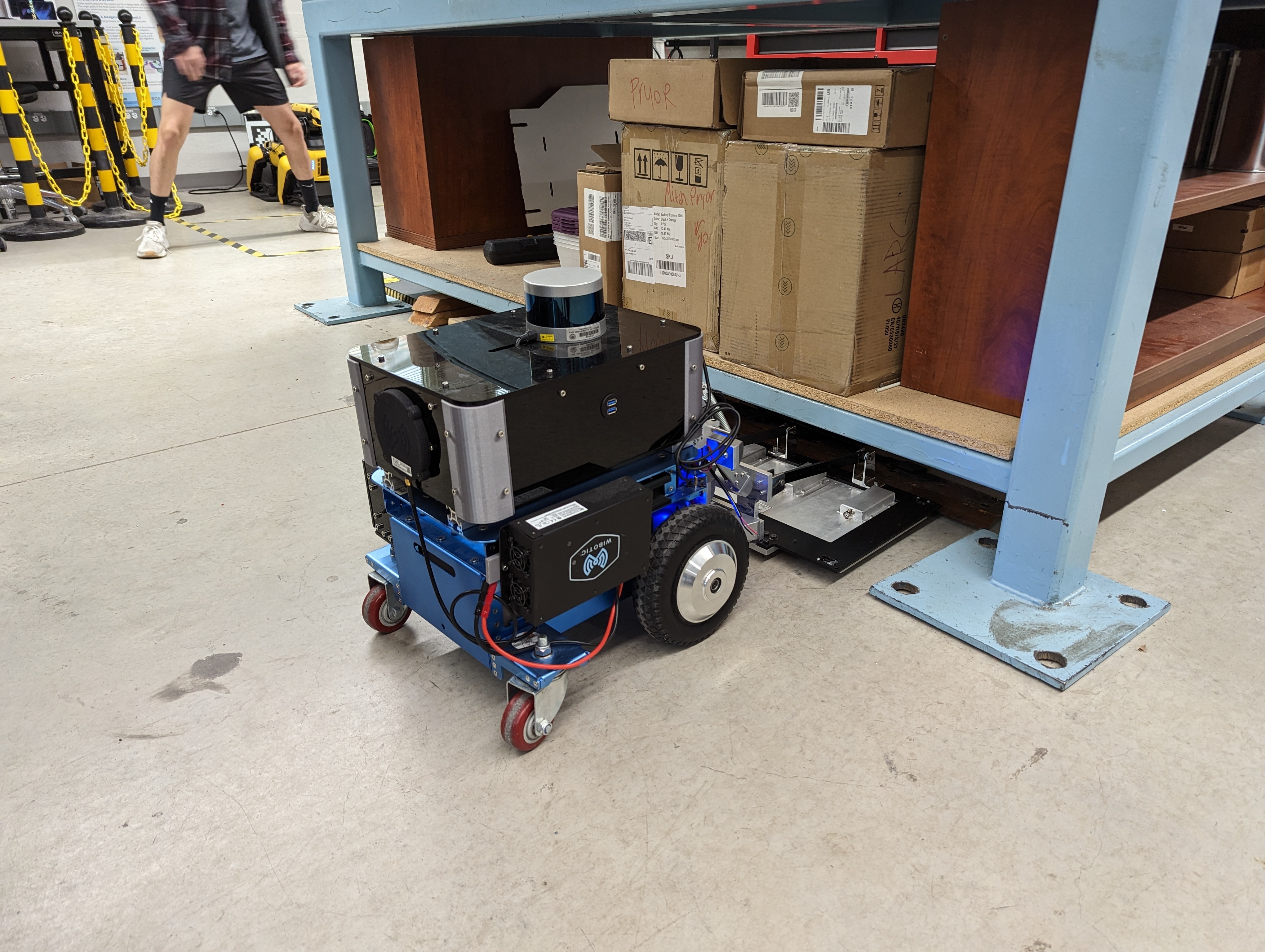}
        \caption{\centering Safe Navigation with Shared Map} 
        \label{fig:minibot_small_obstacle:b} 
    \end{subfigure}

    \caption{\centering Minibot is able to detect hidden obstacles under a table, allowing Magni to safely insert the sensor into the available free space}
    \label{fig:minibot_small_obstacle}
\end{figure}

Another aspect that has not been fully addressed is the ability of the Magni system to collaborate using 3D maps that enable the placement of the sensor under all possible tables [Figure \ref{fig:minibot_small_obstacle:b}]. At this time, the Magni system is able to perform such actions, but it does not yet utilize the highly-detailed maps generated  by the Minibot. This sort of multi-robot integration is expected to improve the overall completeness of the survey.

\section{Conclusion}
\label{sec:conclusion}

In this paper, we document the development of a team of heterogeneous robots that can complete a comprehensive radiation survey in an active nuclear facility. The surveys are routinely performed over extended durations. Previous efforts at the University of Texas at Austin and elsewhere have well-established the ability to complete beta/gamma radiation surveys, and now UT Austin has also developed methods to survey for alpha radiation.

While these efforts give us confidence that the dangerous and dull task of radiation surveys \textit{can} be done by robots, there are significant obstacles to overcome before they \textit{will} be used for this purpose.  Here we presented core capabilities that are necessary for real-world deployment and provide storyboards that enable us to fully understand the complexity of the task still before us. The storyboards provide additional insights into likely challenges with high-level task planning and cybersecurity which includes ongoing plans to port all RRSS components to version 2 of the Robot Operating System (ROS2). 






\bibliographystyle{plain}
\bibliography{References}

\end{document}